# Flame quality monitoring of flare stack based on deep visual features

Xing Mu*
University of Electronic Science and Technology of China
Chengdu,China
2021090916010@std.uestc.edu.cn

## ABSTRACT

Flare stacks play an important role in the treatment of waste gas and waste materials in petroleum fossil energy plants. Monitoring the efficiency of flame combustion is of great significance for environmental protection. The traditional method of monitoring with sensors is not only expensive, but also easily damaged in harsh combustion environments. In this paper, we propose to monitor the quality of flames using only visual features, including the area ratio of flame to smoke, RGB information of flames, angle of flames and other features. Comprehensive use of image segmentation, target detection, target tracking, principal component analysis, GPT-4 and other methods or tools to complete this task. In the end, real-time monitoring of the picture can be achieved, and when the combustion efficiency is low, measures such as adjusting the ratio of air and waste can be taken in time. As far as we know, the method of this paper is relatively innovative and has industrial production value.

## CCS CONCEPTS

•Computing methodologies~Artificial intelligence~Computer vision~Computer vision tasks~Scene anomaly detection

## 1. INTRODUCTION

Petroleum fossil energy has been used in all aspects of our lives. In actual industrial production, the flare tower is an important and irreplaceable safety device. Various waste gases and waste liquids that are difficult to recycle or temporarily unprocessed will be introduced into the flare tower[1]. In the flare tower, these substances will be burned at high temperatures and converted into relatively safe carbon dioxide and water vapor emissions. Therefore, monitoring the combustion efficiency of the flame is of great significance. According to the combustion performance of the flame, we can artificially adjust the supply ratio of fuel and air to make the combustion complete, thereby reducing the impact on the environment.

In the traditional monitoring process, sensors and manual troubleshooting are commonly used methods. However, these two methods have obvious disadvantages. Sensors are mainly used to measure gas composition, but they are not only expensive, but also easily damaged in harsh environments with high temperatures[2]. Manual troubleshooting is dangerous and not sustainable.

Therefore, we proposed a novel flame monitoring method based on visual features, combined with cutting-edge deep learning and other technologies. This method is not only low-cost and can be monitored in real time, but also very effective. We first use the existing data set to train the target detection model, which can realize the detection of flames and smoke, and then complete the segmentation of flames and smoke on this basis, and calculate the area ratio of flames and smoke. We can also calculate the average value of the RGB value of the flame area. The main angle of the flame fluttering (used to judge the wind speed). Then use the principal component analysis method to convert the three sub-variables into two sub-variables (convenient for two-dimensional plane display). After that, we train the data of these two sub-variables. During training, we use GPT-4 to identify whether the combustion efficiency is high, and use the three components as text prompts, taking advantage of the powerful large language model capabilities of GPT-4[3], as shown in Figure 1. Finally, we also expanded the use of tracking algorithms from single torch monitoring to multi-torch monitoring to improve efficiency.

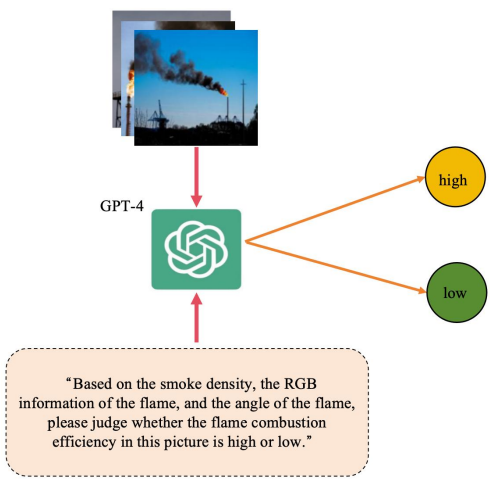

Figure 1: GPT4 preliminary classification based on text prompt.

## 2. RELATED WORKS

### 2.1 YOLOv8 object detection model

YOLOv8 (You Only Look Once, Version 8) is the latest generation of object detection models in the YOLO series[4]. YOLO series models are well-known for their real-time and high efficiency, and are widely used in object detection and image recognition tasks. YOLOv8 uses a deeper convolutional neural network and more feature fusion strategies to better capture details and contextual information in the image. This makes the model perform better when dealing with complex scenes and small object detection. YOLOv8 also introduces new data enhancement techniques and loss functions to further improve the generalization and robustness of the model.

### 2.2 Segment Anything Model

SAM (Segment Anything Model) is a general image segmentation model designed to solve various image segmentation tasks[5]. It is able to segment any type of image without specific training data. It is versatile and can handle a variety of image segmentation tasks, whether it is semantic segmentation, instance segmentation, or panoptic segmentation. It does not need to be retrained for a specific dataset and has strong generalization capabilities. SAM uses advanced neural network architectures such as convolutional neural networks (CNN) and vision transformers. SAM supports users to interact with the model and quickly obtain accurate segmentation results through simple clicks, as shown in Figure 2. This allows users to easily adjust and optimize segmentation results.

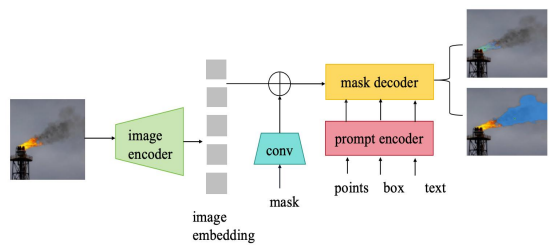

Figure 2:SAM model structure diagram display.

## 2.3 Principal Component Analysis

Principal Component Analysis (PCA) is a commonly used dimensionality reduction technique, mainly used for data preprocessing and feature extraction[6]. It converts the original high-dimensional data into low-dimensional data while trying to retain the main information and structure of the data. The core idea of PCA is to project the original data into a new coordinate system through linear transformation, so that the dimensions of the new coordinate system (i.e., principal components) are orthogonal to each other and sorted from large to small according to the data variance. The first principal component has the largest variance, the second principal component has the second largest variance, and so on. In this way, PCA can reduce the dimensionality of the data while retaining as much data information as possible, as shown in Figure 3.

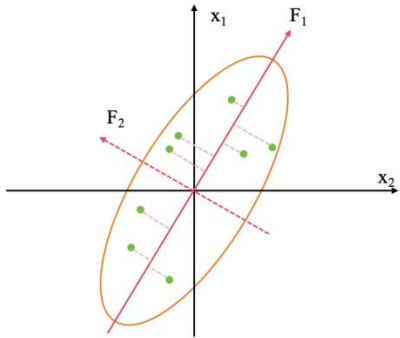

Figure 3: The process of reducing two-dimensional data to one dimension using principal component analysis.

## 2.4 Simple online and realtime tracking

The core idea of SORT is to track multiple targets in real time in a video through a simple linear prediction and data association method[7]. It uses a Kalman filter for state estimation and a Hungarian algorithm for data association to match detection boxes with existing trajectories.

Kalman filtering is a recursive algorithm for estimating the state of a dynamic system, which is widely used in navigation, tracking, control and other fields[8]. It was proposed by Rudolf E. Kálmán in 1960 and combines the mathematical model of the system with noisy observation data to estimate the system state in a way of minimum mean square error.

Status prediction:
$$\hat{x}_{k|k-1} = F_k \hat{x}_{k-1|k-1} + B_k u_k \quad (1)$$

Covariance prediction:
$$P_{k|k-1} = F_k P_{k-1|k-1} F_k^T + Q \quad (2)$$

Calculate Kalman gain:
$$K_k = P_{k|k-1}(H_k P_{k|k-1} H_k^T + R_k)^{-1} \quad (3)$$

State update:
$$\hat{x}_{k|k} = \hat{x}_{k|k-1} + K_k(z_k - H_k \hat{x}_{k|k-1}) \quad (4)$$

Covariance update:
$$P_{k|k} = (I - K_k H_k) P_{k|k-1} \quad (5)$$

The Hungarian algorithm in the SORT (Simple Online and Realtime Tracking) tracking algorithm is an optimization algorithm used to solve the matching problem[9]. Specifically, the Hungarian algorithm is used to solve the maximum weight matching problem in a two-dimensional bipartite graph. In the SORT tracking algorithm, it is used to solve the optimal allocation problem between the target detection box and the tracker.

## 3. METHODOLOGY

In this section, we will demonstrate in detail the entire process of monitoring flame combustion efficiency, including target detection, flame-smoke segmentation, flame drift angle calculation, multi-target flame tracking, etc.

### 3.1 YOLOv8 object detection flame and smoke

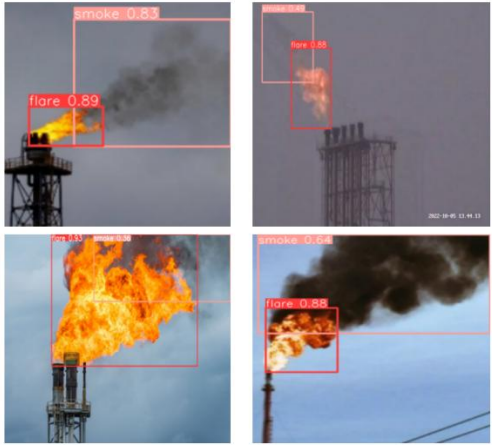

Figure 4:Results of flame and smoke target detection.

We use the YOLOv8 model to train an existing dataset and the results are good on the test set, as shown in Figure 4. Our dataset comes from Abu Dhabi National Oil Company. (ADNOC)

### 3.2 Segmentation of flame and smoke using SAM

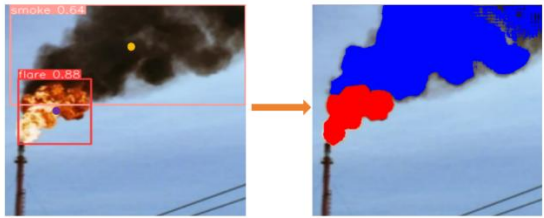

Figure 5:The result of segmentation using the points prompt in SAM in the target detection box.

We use the target detection box and select the midpoint of the detection box (x_center, y_center) as the prompt for SAM input, and obtain a relatively accurate segmentation of flame and smoke, laying the foundation for calculating the ratio of smoke to flame area, as shown in Figure 5. The biggest advantage of this method is that it uses the pre-trained SAM file, and there is no need to train a segmentation model separately, which saves training resources.

### 3.3 Flame Angle Calculation

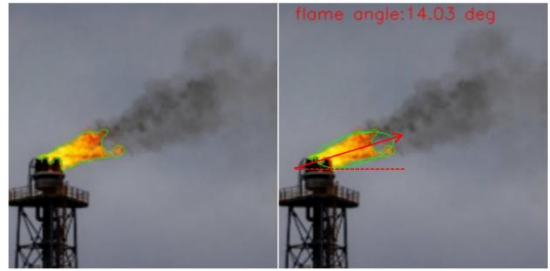

Figure 6:Use an ellipse to fit the flame boundary, and use the major axis direction of the ellipse to approximate the flame direction.

First, the segmentation boundary is obtained from the segmented flame, and then an ellipse is used to fit the boundary. The angle of the line connecting the two ends of the major axis of the ellipse can be approximately regarded as the fluttering angle of the flame, as shown in Figure 6.

### 3.4 SORT algorithm tracks different flames

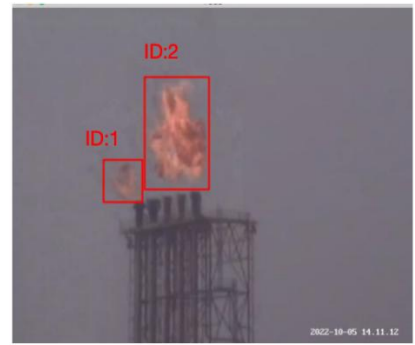

Figure 7:Results of SORT algorithm tracking multiple flames.

When the flare tower has multiple nozzles, we can use the SORT tracking algorithm to independently track different flames, as shown in Figure 7. According to the combustion conditions of different flames, we can specifically adjust the feeding chambers corresponding to different flare heads.

3.5 GPT-4 assists classification

We obtained the numerical smoke-flame ratio, flame RGB average, and flame drift angle of the training data from visual features, but it is difficult to manually assign training labels (whether the combustion efficiency is high). Therefore, we first used GPT-4 to make a preliminary judgment on the combustion efficiency of the flame, and then screened it manually.

3.6 RGB information calculation

We all know that the color of the flame is also related to the combustion efficiency. The blue flame has the highest combustion efficiency, followed by yellow, and red has the lowest. Based on experience and theoretical knowledge, we define an indicator to represent combustion efficiency[10]. We define $w_1$, $w_2$, and $w_3$ as the weights of the blue, yellow, and red components. Their values are 0.7, 0.5 and 0.3 respectively. We first calculate the ratio of the average values of the blue, yellow, and red components of the segmented flame.

$r_1 = V_{avg\_blue}/(V_{avg\_blue} + V_{avg\_yellow} + V_{avg\_red})$ (6)

$r_2 = V_{avg\_green} + V_{avg\_red}/2*(V_{avg\_blue} + V_{avg\_yellow} + V_{avg\_red})$ (7)

$r_3 = V_{avg\_red}/(V_{avg\_blue} + V_{avg\_yellow} + V_{avg\_red})$ (8)

After obtaining the proportion of each component, weighted summation is performed.

$E = w_1 r_1 + w_2 r_2 + w_3 r_3$ (9)

## 4. RESULTS

### 4.1 Presentation of training data

Table 1: Original display of each feature attribute data.

| | smoke_flame ratio | RGB index E | flame angle(°) | label |
|---|---|---|---|---|
| flame1 | 0.22 | 0.62 | 52 | high |
| flame2 | 0.14 | 0.56 | 43 | high |
| flame3 | 0.32 | 0.42 | 23 | high |
| flame4 | 0.40 | 0.36 | 31 | low |
| flame5 | 0.24 | 0.51 | 72 | high |
| flame6 | 0.51 | 0.31 | 34 | low |
| flame7 | 0.62 | 0.24 | 25 | low |
| flame8 | 1.72 | 0.21 | 19 | low |
| flame9 | 2.42 | 0.15 | 12 | low |
| …… | | | | …… |

The data is calculated by the method we mentioned earlier, and its labels are given by GPT4, as shown in Table 1. Next, we standardize the data of each attribute to reduce the differences in distribution between data. Normalize by subtracting the mean and dividing by the variance.

$$X = \frac{x-\mu}{\sigma} \quad (10)$$

Next we calculate the covariance matrix between the three attributes.

$$Cov(X_i, X_j) = \frac{1}{n-1}\sum_{k=1}^{n}(X_{ki} - \bar{X}_i)(X_{kj} - \bar{X}_j) \quad (11)$$

We then calculate the characteristic value and eigenvector.

$$\sum \alpha = \lambda \alpha \quad (12)$$

We select the eigenvectors corresponding to the two largest eigenvalues as projection vectors. We project the standardized data through the projection vectors to obtain the values of the two principal components, as shown in Table 2, Figure 8.

Table 2:Standardize first, then PCA dimension reduction result.

|  | PC1 | PC2 | label |
|---|---|---|---|
| flame1 | -1.89 | 0.21 | high |
| flame2 | -1.43 | -0.22 | high |
| flame3 | -0.11 | -0.85 | high |
| flame4 | -0.07 | -0.45 | low |
| flame5 | -2.10 | 1.04 | high |
| flame6 | 0.10 | -0.23 | low |
| flame7 | 0.74 | -0.48 | low |
| flame8 | 1.89 | 0.30 | low |
| flame9 | 2.88 | 0.68 | low |
| …… |  |  | …… |

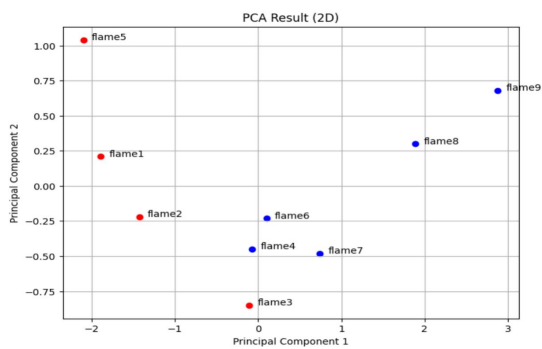

Figure 8:Visual distribution of some training data.

### 4.2 Choice of binary classification model

We use SVM[11], MLP[12], Logic[13], KNN[14] and other classification models to fit the training data, and compare the accuracy on the test data, as shown in Figure 9.

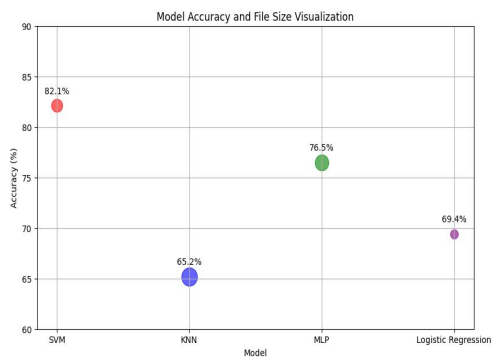

Figure 9:Comparison of the accuracy of different classification models. Points of different sizes represent the size of model parameters.

## 5. CONCLUSION

Our work solves a difficult problem in the petrochemical industry, reducing environmental pollution while ensuring production capacity, and can complete this work at a lower cost. Our work comprehensively utilizes relevant knowledge of computer vision, including target detection, target tracking, and image segmentation. We also use mathematical statistical methods such as principal component analysis to standardize and reduce the dimension of data converted from images or video frames for visualization. We comprehensively use multiple factors to judge the efficiency of flame combustion, use GPT-4 to preliminarily label the training images, and then use the classification model to fit. Finally, we achieved good results on the test set, indicating that this method is relatively stable. Finally, we achieved good results on the test set, indicating that this method is relatively stable.

### ACKNOWLEDGEMENT

I am very grateful to Professor Naoufel Werghi for his guidance and his laboratory for providing relevant datasets.